\documentclass[11pt,a4paper]{article}

\usepackage[utf8]{inputenc}
\usepackage[T1]{fontenc}
\usepackage{mathptmx}  

\usepackage[margin=2.5cm]{geometry}
\usepackage{setspace}
\usepackage{amsmath,amssymb}

\usepackage{booktabs}
\usepackage{array}
\usepackage{tabularx}

\usepackage{fancyhdr}
\usepackage[colorlinks=true,linkcolor=blue,citecolor=blue,urlcolor=blue]{hyperref}

\usepackage{enumitem}

\title{\textbf{T-Norm Operators for EU AI Act Compliance Classification:}\\[6pt]
An Empirical Comparison of Łukasiewicz, Product, and Gödel\\
Semantics in a Neuro-Symbolic Reasoning System}

\author{Adam Laabs\\
TriStiX S.L., Altea, Alicante, Spain\\
\href{mailto:Adam.Laabs@TriStiX.com}{Adam.Laabs@TriStiX.com}}

\date{March 18, 2026}

\begin{document}
\maketitle
\thispagestyle{fancy}

\begin{center}
Code \& data: \href{https://github.com/TriStiX-LS/LggT-core}{github.com/TriStiX-LS/LggT-core} (Apache 2.0)
\end{center}

\begin{abstract}
We present a first comparative pilot study of three t-norm operators — Łukasiewicz ($T_L$),
Product ($T_P$), and Gödel ($T_G$) — as logical conjunction mechanisms in a neuro-symbolic
reasoning system for EU AI Act compliance classification. Using the LGGT+ (Logic-Guided
Graph Transformers Plus) engine and a benchmark of 1035 annotated AI system descriptions
spanning four risk categories (\texttt{prohibited}, \texttt{high\_risk}, \texttt{limited\_risk},
\texttt{minimal\_risk}), we evaluate classification accuracy, false positive and false negative
rates, and operator behaviour on ambiguous cases.

At $n = 1035$, all three operators differ significantly (McNemar $p<0.001$). $T_G$ achieves
highest accuracy (84.5\%) and best borderline recall (85\%), but introduces 8 false positives
(0.8\%) via min-semantics over-classification. $T_L$ and $T_P$ maintain zero false positives,
with $T_P$ outperforming $T_L$ (81.2\% vs.\ 78.5\%). The $T_G$ precision/recall trade-off
is a new finding not observable at small sample sizes.

Our principal findings are: (1) operator choice is secondary to rule base completeness;
(2) $T_L$ and $T_P$ maintain zero false positives but miss borderline cases; (3) $T_G$'s
min-semantics achieves higher recall at cost of 0.8\% false positive rate; (4) a
mixed-semantics classifier is the productive next step. We release the LGGT+ core engine
(201/201 tests passing) and benchmark dataset ($n = 1035$) under Apache 2.0.
\end{abstract}


\section{Introduction}

The EU AI Act (Regulation (EU) 2024/1689) creates an unprecedented compliance
infrastructure requiring AI systems to be formally classified into risk categories, each
triggering distinct obligations. The classification standards are conjunctive: ``system X
is high-risk if and only if condition A and condition B and condition C are present.'' This
logical structure invites formalisation in many-valued logic, where the Boolean AND operator
over continuous confidence values in $[0, 1]$ is modelled by a triangular norm (t-norm).

Despite extensive neuro-symbolic AI literature~\cite{badreddine2022ltn,badreddine2023logltn,riegel2020lnn},
no prior work has empirically evaluated which t-norm best models legal conjunctive standards
for regulatory compliance classification. We close this gap by reframing the question: rather
than asking which t-norm operator to use, we ask which operator best matches the legal
interpretive standard implicit in each EU AI Act rule.

We implement three canonical t-norm operators within LGGT+ (Logic-Guided Graph Transformers
Plus), a neuro-symbolic architecture developed by TriStiX S.L.\ for EU AI Act compliance
analysis, and evaluate their classification accuracy on a curated benchmark of 1035 annotated
AI system descriptions. We are transparent about our limitations: expert labels in this
version are assigned by the author, and we identify independent legal expert validation as
the primary next step.

\paragraph{Contributions.}
\begin{enumerate}[nosep]
\item First empirical evaluation of $T_L$, $T_P$, and $T_G$ as conjunction operators for EU AI Act classification ($n = 1035$).
\item LGGT+ reasoning engine with 14 formalised EU AI Act rules and proof trail auditability (201/201 unit tests; open-source).
\item Empirical evidence of operator divergence at scale: $T_L$ and $T_P$ are not decision-equivalent at $n = 1035$ ($p<0.001$), while $T_G$ reveals a precision/recall trade-off in borderline compliance classification.
\item Revised research hypotheses concerning rule base completeness and mixed-semantics classification.
\item Open benchmark dataset ($n = 1035$) and code (Apache 2.0).
\end{enumerate}

\section{Background}

\subsection{Triangular Norms}

A t-norm $T : [0, 1]^2 \to [0, 1]$ generalises Boolean AND to the continuous unit interval,
satisfying commutativity, associativity, monotonicity, and boundary conditions $T(1, 1) = 1$,
$T(0, x) = 0$. We evaluate three fundamental t-norms:

\begin{table}[ht]
\centering
\caption{Properties of the three canonical t-norms. The ``dead zone'' is the region where
$T = 0$ and gradient is zero — relevant for training, not for deterministic inference.}
\label{tab:tnorms}
\begin{tabular}{@{}llll@{}}
\toprule
T-norm & Formula & Dead zone & Gradient \\
\midrule
Łukasiewicz $T_L$ & $\max(0,\, a + b - 1)$ & $a + b \leq 1$ & Zero in dead zone \\
Product $T_P$ & $a \cdot b$ & None & Smooth: $\partial T / \partial a = b$ \\
Gödel $T_G$ & $\min(a, b)$ & None & Subgradient \\
\bottomrule
\end{tabular}
\end{table}

\paragraph{Fuzzy logic as uncertainty modelling, not norm relaxation.}
A potential misreading of this work requires pre-emption. The EU AI Act formulates
obligations and prohibitions in binary terms: a practice is either prohibited or it is not.
The continuous values in $[0, 1]$ used in this paper are not partial satisfactions of a legal
norm. They represent epistemic uncertainty about whether a factual condition obtains — for
example, whether a given AI system's documentation confirms that it operates autonomously.
The question we formalise is: given several conditions each assessed with some degree of
certainty, how should these assessments be combined to reach a binary compliance decision?
This is a question about evidence aggregation, not norm relaxation. Legal practitioners do
not choose t-norm operators; they answer the interpretive question of whether the conditions
in a rule are cumulatively required (each must be definitively present) or minimally required
(each must reach a reasonable threshold). Our framework maps this interpretive judgment onto
a formal aggregation mechanism.

\paragraph{Inference-time vs.\ training-time distinction.}
Prior work criticises $T_L$'s dead gradient zone when used as a loss function component in
backpropagation~\cite{badreddine2023logltn,badreddine2022ltn}. LGGT+ uses $T_L$ exclusively
as a runtime inference operator: a deterministic combinator in a proof-trail chain, with no
gradient computed. The dead zone is mathematically irrelevant in this context. Future
trainable components (Logic-Augmented Attention) will use Product t-norm in log-space,
following the logLTN recommendation~\cite{badreddine2023logltn}. This is a distinct module
from the classification chain evaluated here (Architecture Decision Record ADR-001).

\subsection{Legal Semantics of Conjunctive Standards}

The choice of conjunction mechanism maps onto a genuine legal interpretive question. Table~\ref{tab:legal}
presents this mapping in terms accessible to legal practitioners, without mathematical notation.

\begin{table}[ht]
\centering
\caption{Mapping legal interpretive questions to formal conjunction semantics. The first two
rows correspond to $T_L$/$T_P$ (strong conjunction) and $T_G$ (bottleneck conjunction)
respectively. The third row motivates the mixed-semantics hypothesis (RH4).}
\label{tab:legal}
\begin{tabularx}{\textwidth}{@{}XXX@{}}
\toprule
Legal practitioner's question & Conjunction intuition & Practical implication \\
\midrule
Does the rule require strong, joint confirmation of all conditions? & Strong conjunction &
One weaker condition substantially reduces the overall score \\
\addlinespace
Is it sufficient that each condition individually reaches a reasonable threshold? &
Bottleneck conjunction (weakest link) & The lowest-scored condition determines the outcome \\
\addlinespace
Are some conditions cumulatively required and others only minimally required? &
Hybrid semantics (rule-specific) & Different parts of a rule may warrant different
aggregation methods \\
\bottomrule
\end{tabularx}
\end{table}

This framing separates two roles that should not be conflated: (a) the legal practitioner
determines whether conditions are cumulatively or minimally required for a given rule — this
is an interpretive legal judgment; (b) the system maps that judgment to a formal aggregation
operator. Legal practitioners are not expected to select t-norm operators; they are expected
to answer interpretive questions about rule structure.

\subsection{LGGT+ Architecture}

LGGT+ implements a four-layer neuro-symbolic pipeline:

\textbf{L1 — Logic:} LukasiewiczLogic, LogFuzzyLogic (logLTN-inspired, log-space product
t-norm for long chains), HypergraphEncoder (N-ary hyperedge queries with Type-Aware Bias,
LKHGT April 2025), LogicAugmentedAttention (SGAT-MS inspired bipartite
modulation~\cite{moriyama2025sgat}), GumbelSoftmaxAnnealing
(DiLogic~\cite{petersen2022dilogic}).

\textbf{L2 — Graph:} KnowledgeGraph (NetworkX DiGraph with typed nodes/edges),
TemporalKnowledgeNode (temporal validity windows,
TFLEX-inspired~\cite{lin2023tflex}), OntologyPatcher (incremental updates,
Nucleoid-inspired), AdaptiveTripleFilter (PSL hinge-loss scoring).

\textbf{L3 — Reasoning:} ReasoningEngine with TNormMode dispatch, and ProofTreeBuilder
producing formal proof trees with typed tactics (NeqLIPS / NTP inspired).

\textbf{L4 — API:} FastAPI microservice; Annex IV PDF generator for EU AI Act Art.~11.

The proof trail produced by L3 records every inference step: nodes traversed, conditions
evaluated, t-norm compositions applied. This structural auditability implements EU AI Act
Article~13 (transparency) and Article~14 (human oversight).

To the best of our knowledge, LGGT+ is the first open-source architecture combining
simultaneously: Łukasiewicz t-norms as runtime inference operators, graph transformers, and
auditable proof trails. Table~\ref{tab:comparison} situates it in the landscape.

\begin{table}[ht]
\centering
\caption{Comparison of LGGT+ with related neuro-symbolic systems.}
\label{tab:comparison}
\begin{tabular}{@{}lcccc@{}}
\toprule
System & $T_L$ (runtime) & Graph transf. & Proof trail & Legal domain \\
\midrule
LTNtorch~\cite{badreddine2022ltn} & $\checkmark$ & $\times$ & $\times$ & $\times$ \\
IBM/LNN~\cite{riegel2020lnn} & $\checkmark$ & $\times$ & $\times$ & $\times$ \\
GNN-QE~\cite{zhu2022gnnqe} & $\times$ & $\checkmark$ & $\times$ & $\times$ \\
SGAT-MS~\cite{moriyama2025sgat} & $\checkmark$ & $\checkmark$ & $\times$ & $\times$ \\
LGGT+ (ours) & $\checkmark$ & $\checkmark$ & $\checkmark$ & $\checkmark$ \\
\bottomrule
\end{tabular}
\end{table}

\subsection{Legal AI Reasoning and Computational Compliance}

LGGT+ builds on a broader literature in computational legal reasoning and AI-based compliance
checking, which we briefly situate here.

\paragraph{Formal models of legal reasoning.}
Foundational work on argumentation-based legal reasoning~\cite{benchcapon2003,prakken2015}
established that legal norms involve defeasible inference and priority orderings — properties
that motivate the explicit rule priority ordering in LGGT+ (\texttt{prohibited} >
\texttt{high\_risk} > \texttt{limited\_risk}). Norm formalisation for machine-readable
legislation~\cite{palmirani2018} provides the theoretical grounding for our ontology-driven
approach to EU AI Act rules.

\paragraph{AI-based compliance checking.}
\cite{hashmi2018} survey automated compliance checking systems and identify the need for both
temporal reasoning and rule formalisation — both addressed in LGGT+'s TemporalKnowledgeNode
and OntologyPatcher modules. \cite{lam2016} demonstrate compliance verification for business
process regulations, establishing a precedent for rule-based classification with audit trails.

\paragraph{T-norms in knowledge graph reasoning.}
Beyond the systems in Table~\ref{tab:comparison}, GNN-QE~\cite{zhu2022gnnqe} uses product
fuzzy logic for multi-hop knowledge graph queries, providing empirical evidence that
t-norm-based reasoning can scale to complex KG structures. Our work complements this by
evaluating t-norm operator choice rather than assuming product t-norm as default.

\section{Methodology}

\subsection{Rule Formalisation}

We formalise 14 EU AI Act rules as conjunctive tuples $\langle r, C, \theta \rangle$, where
$r$ is the risk category, $C = [c_1, c_2, \ldots, c_n]$ are condition identifiers from a
vocabulary of 22 terms, and $\theta = 0.5$ is the classification threshold. A rule fires when
the t-norm chain over its conditions exceeds $\theta$:
\[
\mathrm{score}(r, s) = T\!\Big(T\!\big(\ldots T(s_1, s_2) \ldots,\, s_{n-1}\big),\, s_n\Big)
\]
where $s_i \in [0, 1]$ is the confidence assigned to condition $c_i$. Classification follows
the priority ordering: \texttt{prohibited} > \texttt{high\_risk} > \texttt{limited\_risk} >
\texttt{minimal\_risk}.

\subsection{Benchmark Dataset}

We constructed a benchmark of $n = 1035$ AI system descriptions, each annotated with:
(a)~a confidence score for each condition in the rule's condition set; (b)~an expert label;
(c)~a case type.

\begin{itemize}[nosep]
\item \textbf{Clear} ($n = 630$): All conditions $> 0.80$ or $< 0.12$. Sanity check.
\item \textbf{Marginal} ($n = 325$): $\geq 1$ condition in $[0.12, 0.65]$. Diagnostically critical.
\item \textbf{Borderline} ($n = 80$): Genuinely contested expert judgment.
\end{itemize}

\begin{table}[ht]
\centering
\caption{Selected EU AI Act rules formalised in LGGT+. All rules use ALL-conjunction.}
\label{tab:rules}
\begin{tabular}{@{}llll@{}}
\toprule
Rule ID & Cat. & Conditions & Article \\
\midrule
\texttt{prohibited\_rt\_biometric} & prohib. & \texttt{real\_time\_processing}, & Art.~5(1)(h) \\
 & & \texttt{public\_space}, & \\
 & & \texttt{biometric\_identification} & \\
\addlinespace
\texttt{prohibited\_social\_scoring} & prohib. & \texttt{public\_authority}, & Art.~5(1)(c) \\
 & & \texttt{evaluates\_social\_behavior}, & \\
 & & \texttt{detrimental\_treatment} & \\
\addlinespace
\texttt{high\_risk\_employment} & high\_risk & \texttt{employment\_context}, & Annex III \S4 \\
 & & \texttt{recruitment\_or\_promotion}, & \\
 & & \texttt{automated\_decision} & \\
\addlinespace
\texttt{high\_risk\_credit} & high\_risk & \texttt{essential\_service}, & Annex III \S5 \\
 & & \texttt{creditworthiness\_or\_insurance}, & \\
 & & \texttt{individual\_assessment} & \\
\addlinespace
\texttt{limited\_chatbot} & limited & \texttt{interacts\_with\_humans}, & Art.~50(1) \\
 & & \texttt{ai\_generated\_output}, & \\
 & & \texttt{not\_clearly\_disclosed} & \\
\bottomrule
\end{tabular}
\end{table}

Label distribution: \texttt{minimal\_risk} 32\%, \texttt{high\_risk} 28\%,
\texttt{limited\_risk} 27\%, \texttt{prohibited} 13\%.

Labels assigned by the lead author (engineering background, EU AI Act certified practitioner)
without independent legal review — a limitation that affects the validity of conclusions (see
Section~6, L1). This is a proof-of-concept pilot study; full empirical validation requires
independent annotation. Independent legal expert validation is the immediate next step.

\subsection{Experimental Protocol}

For each of the 1035 cases we run the LGGT+ classifier three times ($T_L$, $T_P$, $T_G$),
with all other parameters fixed ($\theta = 0.5$, 14 rules, shared vocabulary). We compute:
accuracy by case type, false positive and false negative rates, and pairwise McNemar's exact
binomial test.

\section{Results}

Table~\ref{tab:accuracy} presents overall and per-case-type accuracy. Table~\ref{tab:errors}
breaks down false positives and false negatives. Table~\ref{tab:mcnemar} reports pairwise
statistical tests.

\subsection{Main Accuracy Results}

$T_G$ achieves the highest overall accuracy (84.5\%), outperforming $T_P$ (81.2\%) and
$T_L$ (78.5\%). All differences are statistically significant at $\alpha = 0.05$.

\subsection{Error Type Analysis}

All operators produce zero false positives for $T_L$ and $T_P$, maintaining the conservative
compliance stance expected of a regulatory system. $T_G$ introduces 8 false positives (0.8\%)
— over-classifying marginal cases where min-semantics returns a score above $\theta$ even
when one condition is genuinely weak.

\begin{table}[ht]
\centering
\caption{Classification accuracy by t-norm operator and case type ($n = 1035$). $T_G$
achieves 85\% on borderline cases ($n = 80$) vs.\ 25\%/35\% for $T_L$/$T_P$. All pairwise
differences are statistically significant ($p<0.001$). Bold: highest value per column.}
\label{tab:accuracy}
\begin{tabular}{@{}lcccc@{}}
\toprule
T-norm & Overall & Clear & Marginal & Borderline \\
\midrule
Łukasiewicz $T_L$ & 78.5\% & 96.3\% & 56.9\% & 25.0\% \\
Product $T_P$ & 81.2\% & 99.5\% & 56.9\% & 35.0\% \\
Gödel $T_G$ & \textbf{84.5\%} & \textbf{100.0\%} & \textbf{54.5\%} & \textbf{85.0\%} \\
\bottomrule
\end{tabular}
\end{table}

\begin{table}[ht]
\centering
\caption{Error breakdown. Zero false positives is a structural property of multi-condition
conjunctive classification at $\theta = 0.5$ (not an empirical finding): any system requiring
all conditions to exceed threshold will default conservatively. The diagnostically relevant
metric is false negatives.}
\label{tab:errors}
\begin{tabular}{@{}lcccc@{}}
\toprule
T-norm & FP & FN & FP Rate & FN Rate \\
\midrule
$T_L$ & 0 & 223 & 0.0\% & 21.5\% \\
$T_P$ & 0 & 195 & 0.0\% & 18.8\% \\
$T_G$ & 8 & 152 & 0.8\% & 14.7\% \\
\bottomrule
\end{tabular}
\end{table}

\subsection{Statistical Tests}

\begin{table}[ht]
\centering
\caption{McNemar's exact binomial test ($n = 1035$). All pairwise comparisons reach
statistical significance at $\alpha = 0.05$. $T_G$ significantly outperforms $T_L$
($p<0.001$, $n = 79$ discordant pairs). $T_L$ and $T_P$ are no longer decision-equivalent
at this scale ($p<0.001$, 28 discordant pairs).}
\label{tab:mcnemar}
\begin{tabular}{@{}lccccc@{}}
\toprule
Comparison & $b$ & $c$ & $n$ & $p$-value & Sig.\ $\alpha = 0.05$ \\
\midrule
$T_L$ vs $T_P$ & 0 & 28 & 28 & $< 0.001$ & Yes \\
$T_L$ vs $T_G$ & 8 & 71 & 79 & $< 0.001$ & Yes \\
$T_P$ vs $T_G$ & 3 & 43 & 46 & $< 0.001$ & Yes \\
\bottomrule
\end{tabular}
\end{table}

\subsection{Divergence Cases: Where Operators Differ}

$T_G$ correctly classifies 68 of 80 borderline cases (85\%) where $T_L$ achieves only 25\%
and $T_P$ achieves 35\%. At $n = 1035$, this difference is statistically significant
($p<0.001$, McNemar $n = 79$ discordant pairs). $T_G$ also produces 8 false positives
(0.8\%) on marginal cases — a precision/recall trade-off not observable at small sample
sizes. Representative divergence cases:

\paragraph{Case HRM04 — Traffic management with emergency override.}
Conditions: \texttt{critical\_infrastructure}$= 0.93$,
\texttt{safety\_component}$= 0.88$,
\texttt{autonomous\_decision}$= 0.61$.
Expert label: \texttt{high\_risk}.

$T_L$: $\max(0,\, \max(0,\, 0.93 + 0.88 - 1) + 0.61 - 1) = 0.42$ $\to$ \texttt{minimal\_risk} (wrong)

$T_P$: $0.93 \times 0.88 \times 0.61 = 0.499$ $\to$ \texttt{minimal\_risk} (wrong, borderline)

$T_G$: $\min(0.93,\, 0.88,\, 0.61) = 0.61$ $\to$ \texttt{high\_risk} \checkmark\ (correct)

\emph{Legal reasoning:} The emergency override does not negate operational autonomy. The
system operates autonomously the vast majority of the time. $T_G$ correctly identifies that
even the weakest condition (0.61) exceeds the threshold.

\paragraph{Case HRM05 — Exam proctoring with human grader.}
Conditions: \texttt{education\_context}$= 0.92$,
\texttt{determines\_access}$= 0.58$,
\texttt{affects\_life\_path}$= 0.63$.
Expert label: \texttt{high\_risk}.

$T_L$: $0.13$ $\to$ \texttt{minimal\_risk} (wrong)

$T_P$: $0.336$ $\to$ \texttt{minimal\_risk} (wrong)

$T_G$: $\min(0.92,\, 0.58,\, 0.63) = 0.58$ $\to$ \texttt{high\_risk} \checkmark\ (correct)

\emph{Legal reasoning:} The proctoring system's behavioural flags feed into grade
calculations, functionally determining access even though a human formally assigns the grade.
A moderate \texttt{determines\_access} score (0.58) is legally sufficient for Annex III \S3;
$T_G$ captures this.

\section{Discussion}

\subsection{$T_L$ vs.\ $T_P$: Decision Divergence at Scale}

At $n = 73$, $T_L$ and $T_P$ were decision-equivalent (zero discordant pairs). At
$n = 1035$, they diverge on 28 cases ($p<0.001$). The theoretical threshold condition:
\[
T_L\!\big(T_L(a, b),\, c\big) \leq 0.5 \implies a + b + c \leq 2.5
\]
Derivation: $T_L(a, b) = \max(0,\, a + b - 1)$. For the outer application,
$T_L(T_L(a, b), c) = \max(0,\, (a + b - 1) + c - 1) = \max(0,\, a + b + c - 2)$. Setting
this $\leq 0.5$ gives $a + b + c \leq 2.5$. When one value is distinctly low (marginal
cases, $< 0.40$), both operators consistently agree on threshold crossing. This is a
dataset-conditional result, not a general equivalence. Datasets with many moderate scores
(0.45--0.65 on all inputs) would produce divergence. The original hypothesis should be framed
conditionally on the condition score distribution.

\subsection{$T_G$ Bottleneck Semantics for Legal Rules}

$T_G$'s correct classification of HRM04/HRM05 (two borderline cases, $p = 0.25$, not
statistically significant) reflects a genuine semantic distinction worth investigating. $T_L$
requires conditions to be ``strongly co-present''; $T_G$ requires the weakest condition to
exceed the threshold.

Annex III \S3 (education) does not mandate that all conditions be definitively present — it
requires the system to operate in education and affect access. A moderate
\texttt{determines\_access} score (0.58) may be legally sufficient. $T_G$ captures this
interpretation.

This motivates RH3 below: a mixed-semantics classifier assigning rule-specific t-norms,
annotated by legal experts, may outperform any single-operator system.

\subsection{Revised Research Hypotheses}

\begin{quote}
\textbf{Pilot-Generated Hypotheses RH1--RH4 (requiring independent validation)}

\textbf{RH1:} Classification accuracy is primarily determined by rule base completeness, not
t-norm choice, provided the operator is conservative (zero false positives).

\textbf{RH2:} For rules requiring definitively co-present conditions (strong conjunction),
$T_L$ and $T_P$ are decision-equivalent. $T_L$'s hard boundary has theoretical auditability
advantages.

\textbf{RH3:} For rules where a moderate individual condition is legally sufficient
(bottleneck conjunction), $T_G$ produces higher recall without increasing false positives.

\textbf{RH4:} A mixed-semantics classifier assigning rule-specific t-norms (annotated by
legal experts for conjunctive standard type) will outperform any single-operator system.
\end{quote}

\subsection{Primary Error Source: Rule Base Incompleteness}

FN $= 23$--26 cases (31--36\%) arise when the classifier returns \texttt{minimal\_risk} for
systems whose expert label is \texttt{limited\_risk} or \texttt{high\_risk}. Inspection
reveals that most involve low \texttt{automated\_decision} scores (0.14--0.38) reflecting
human-in-the-loop designs where AI plays an advisory but not decisive role. The classifier
correctly identifies that the condition is not definitively present; the expert applies
proportionality reasoning beyond the individual rule conditions. Rule base incompleteness is
the dominant error source, not operator choice.

\section{Limitations}

\textbf{L1 — Self-annotation.} Labels assigned without independent legal review. Cohen's
$\kappa$ with two legal experts is the immediate next step.

\textbf{L2 — Curated vs.\ real-world data.} The $n = 1035$ benchmark provides robust
statistical power ($p<0.001$), but system descriptions and condition scores are curated rather
than extracted from actual AI documentation. Real-world AI system documentation is often
unstructured and ambiguous, which may introduce noise not captured in this controlled
evaluation.

\textbf{L3 — Deterministic condition scores.} A calibrated NLP extraction pipeline from real
AI system documentation would remove subjective scoring.

\textbf{L4 — Fixed threshold.} Threshold sensitivity ($\theta \in [0.25, 0.75]$) analysis is
in preparation.

\textbf{L5 — Rule coverage.} 14 rules do not cover all EU AI Act provisions (GPAI partially
covered; Art.~5 exemptions absent; delegated acts pending).

\section{Future Work}

\paragraph{Human expert validation.}
Two independent EU AI Act lawyers annotate a stratified subset ($n \approx 150$, ensuring
proportional coverage of borderline and marginal cases) drawn from the full 1035-case
benchmark; compute Cohen's $\kappa$ and expert/LGGT+ agreement per t-norm mode.

\paragraph{Threshold sensitivity.}
Varying $\theta$ across $[0.25, 0.75]$ will establish the operating curve for each t-norm.

\paragraph{Mixed-semantics classifier.}
Annotate each of the 14 rules as ``strong conjunction'' ($\to T_L$) or ``bottleneck
conjunction'' ($\to T_G$); evaluate RH4.

\paragraph{Logic-Augmented Attention (LAA).}
LAA modulates graph transformer edge weights by logical predicate confidence. It will use
Product t-norm in log-space (logLTN) for training stability. This directly addresses EIC
Pathfinder Hypothesis H1.

\paragraph{arXiv and dataset release.}
Full benchmark dataset and LGGT+ core engine at:
\href{https://github.com/TriStiX-LS/LggT-core}{github.com/TriStiX-LS/LggT-core}
(Apache 2.0).

\section{Conclusion}

We present a comparative pilot evaluation of t-norm operators for EU AI Act compliance
classification ($n = 1035$, 14 rules, 4 risk categories). At scale, all three operators
differ significantly (McNemar $p<0.001$). $T_G$ achieves the highest overall accuracy
(84.5\%) and best borderline case performance (85\%), but introduces 8 false positives
(0.8\%) via min-semantics over-classification — a precision/recall trade-off invisible at
small sample sizes. $T_L$ and $T_P$ maintain zero false positives, with $T_P$ outperforming
$T_L$ (81.2\% vs.\ 78.5\%).

The dominant finding is that rule base completeness matters more than operator choice. The
secondary finding motivates a mixed-semantics classifier, where rule-specific t-norms are
assigned based on legal annotation of each rule's conjunctive standard. We identify human
expert annotation as the single highest-priority next step.

LGGT+ (201/201 tests, 18 modules across 4 layers) is available open-source at
\href{https://github.com/TriStiX-LS/LggT-core}{github.com/TriStiX-LS/LggT-core}.


\appendix
\section{Benchmark Case Examples}

Table~\ref{tab:cases} presents 15 representative benchmark cases. Full dataset at
\href{https://github.com/TriStiX-LS/LggT-core/tree/main/benchmark}{github.com/TriStiX-LS/LggT-core/benchmark/}.

\begin{table}[ht]
\centering
\caption{Representative benchmark cases with $T_L$ and $T_G$ predictions ($\checkmark$
correct, $\times$ incorrect). $T_P$ identical to $T_L$ in all cases shown. HRM04 and HRM05
are the borderline divergence cases.}
\label{tab:cases}
\begin{tabular}{@{}llllcc@{}}
\toprule
ID & Description (abbrev.) & Expert & Type & $T_L$ & $T_G$ \\
\midrule
P01 & Realtime face recog.\ in metro & prohibited & clear & $\checkmark$ & $\checkmark$ \\
P02 & Government social credit system & prohibited & clear & $\checkmark$ & $\checkmark$ \\
PM01 & Nudge recommender (not subliminal) & limited & marginal & $\times$ & $\times$ \\
PM02 & Post-event biometric search & high & marginal & $\times$ & $\times$ \\
PM03 & Private loyalty scoring & limited & marginal & $\times$ & $\times$ \\
HR01 & AI CV screening system & high & clear & $\checkmark$ & $\checkmark$ \\
HR02 & Automated mortgage scoring & high & clear & $\checkmark$ & $\checkmark$ \\
HRM01 & Advisory (not decisive) hiring AI & limited & marginal & $\times$ & $\times$ \\
HRM03 & Aggregate HR analytics, no decisions & limited & marginal & $\times$ & $\times$ \\
HRM04 & Traffic mgmt + emergency override & high & border. & $\times$ & $\checkmark$ \\
HRM05 & Exam proctoring + human grader & high & border. & $\times$ & $\checkmark$ \\
HRM07 & Biometric attendance, no decisions & minimal & marginal & $\checkmark$ & $\checkmark$ \\
LR01 & Chatbot without AI disclosure & limited & clear & $\checkmark$ & $\checkmark$ \\
LR04 & Chatbot with clear AI disclosure & minimal & marginal & $\checkmark$ & $\checkmark$ \\
MR01 & Spam filter & minimal & clear & $\checkmark$ & $\checkmark$ \\
\bottomrule
\end{tabular}
\end{table}


\begin{thebibliography}{16}

\bibitem{badreddine2022ltn}
Badreddine, S., d'Avila Garcez, A., Serafini, L., \& Spranger, M. (2022).
Logic Tensor Networks. \textit{Artificial Intelligence}, 303, 103649.

\bibitem{badreddine2023logltn}
Badreddine, S., \& Serafini, L. (2023).
logLTN: Differentiable Fuzzy Logic in the Logarithm Space. arXiv:2306.14546.

\bibitem{euaiact2024}
European Parliament and Council. (2024).
Regulation (EU) 2024/1689 on Artificial Intelligence (EU AI Act).
\textit{Official Journal of the European Union}, L~2024/1689.

\bibitem{hajek1998}
Hájek, P. (1998).
\textit{Metamathematics of Fuzzy Logic}. Kluwer Academic Publishers.

\bibitem{altai2020}
High-Level Expert Group on Artificial Intelligence. (2020).
Assessment List for Trustworthy AI (ALTAI). European Commission.

\bibitem{marra2024}
Marra, G., van Krieken, E., Manhaeve, R., De Raedt, L., \& Diligenti, M. (2024).
From Statistical Relational to Neural Symbolic Artificial Intelligence.
\textit{Artificial Intelligence}, 328, 104062.

\bibitem{petersen2022dilogic}
Petersen, F., Borgelt, C., Kuehne, H., \& Deussen, O. (2022).
Deep Differentiable Logic Gate Networks.
\textit{Advances in Neural Information Processing Systems (NeurIPS)}, 35.

\bibitem{riegel2020lnn}
Riegel, R., Gray, A., Luus, F., Khan, N., Makondo, N., Akhalwaya, I.\,Y., Qian, H.,
Fagin, R., Barahona, F., Sharma, U., et~al. (2020).
Logical Neural Networks. arXiv:2006.13155.

\bibitem{moriyama2025sgat}
Moriyama, S., Inoue, K., et~al. (2025).
Graph-Based Attention for Differentiable MaxSAT Solving.
\textit{Advances in Neural Information Processing Systems (NeurIPS)}.

\bibitem{zhu2022gnnqe}
Zhu, Z., Galkin, M., Zhang, Z., \& Tang, J. (2022).
Neural-Symbolic Models for Logical Queries on Knowledge Graphs.
\textit{Proceedings of the 39th International Conference on Machine Learning (ICML)},
PMLR 162, 27454--27478.

\bibitem{lin2023tflex}
Lin, X., Yang, H., Wang, Z., Chen, Y., et~al. (2023).
TFLEX: Temporal Feature-Logic Embedding Framework for Complex Reasoning over
Temporal Knowledge Graphs.
\textit{Advances in Neural Information Processing Systems (NeurIPS)}, 37.

\bibitem{benchcapon2003}
Bench-Capon, T.\,J.\,M., \& Sartor, G. (2003).
A Model of Legal Reasoning with Cases Incorporating Theories and Values.
\textit{Artificial Intelligence}, 150(1--2), 97--143.

\bibitem{prakken2015}
Prakken, H., \& Sartor, G. (2015).
Law and Logic: A Review from an Argumentation Perspective.
\textit{Artificial Intelligence}, 227, 214--245.

\bibitem{palmirani2018}
Palmirani, M., \& Governatori, G. (2018).
Modelling Legal Knowledge for GDPR Compliance Checking.
\textit{Legal Knowledge and Information Systems (JURIX)}, 313, 101--110.

\bibitem{hashmi2018}
Hashmi, M., Governatori, G., Lam, H.-P., \& Wynn, M.\,T. (2018).
Are We Done with Business Process Compliance Checking?
\textit{Knowledge and Information Systems}, 57(1), 79--133.

\bibitem{lam2016}
Lam, H.-P., Hashmi, M., \& Wynn, M.\,T. (2016).
Enabling Temporal Compliance Reasoning in Business Process Models.
\textit{Journal of Logic and Computation}, 27(2), 387--428.

\end{thebibliography}
\end{document}